\pdfoutput=1

\documentclass[11pt]{article}

\usepackage[table,xcdraw]{xcolor}

\usepackage[]{ACL2023}

\usepackage{times}
\usepackage{latexsym}

\usepackage[T1]{fontenc}

\usepackage[utf8]{inputenc}

\usepackage{microtype}

\usepackage{inconsolata}
\newcommand\blfootnote[1]{%
\begingroup
\renewcommand\thefootnote{}\footnote{#1}%
\addtocounter{footnote}{-1}%
\endgroup
}

\usepackage{booktabs} 
\usepackage{tabularx}

\usepackage{colortbl}
\usepackage{tabularx}
\usepackage{booktabs}
\definecolor{lightest}{HTML}{fcfbe1}
\definecolor{light}{HTML}{e7faaa}
\definecolor{medium}{HTML}{a9eb78}
\definecolor{dark}{HTML}{5ec467}
\definecolor{darkest}{HTML}{2bb338}
\definecolor{gray}{HTML}{66638a}

\usepackage[]{graphicx}

\usepackage{amssymb}
\usepackage{pifont}
\newcommand{\cmark}{\ding{51}}%
\newcommand{\xmark}{\ding{55}}%

\usepackage{multirow}
\newcommand{\hD}{\mathcal{D}}
\newcommand{\hQ}{\mathcal{Q}}
\newcommand{\hS}{\mathcal{S}}

\definecolor{codegreen}{rgb}{0,0.6,0}
\definecolor{codegray}{rgb}{0.5,0.5,0.5}
\definecolor{codepurple}{rgb}{0.58,0,0.82}
\definecolor{backcolour}{rgb}{0.95,0.95,0.92}
\usepackage{listings}

\usepackage{listings}
\lstdefinestyle{mystyle}{
  commentstyle=\color{codegreen},
  keywordstyle=\color{magenta},
  numberstyle=\small\color{codegray},
  stringstyle=\color{codepurple},
  basicstyle=\ttfamily\tiny,
  breakatwhitespace=false,         
  breaklines=true,                 
  captionpos=b,                    
  keepspaces=false,                                 
  showspaces=false,                
  showstringspaces=false,
  showtabs=false,                  
  tabsize=2
}

\usepackage{amsmath}
\usepackage{algorithm}
\usepackage{algpseudocode}
\title{MathGenie: Generating Synthetic Data with Question Back-translation for Enhancing Mathematical Reasoning of LLMs}

\author{Zimu Lu$^{*1}$, Aojun Zhou$^{*1}$, Houxing Ren$^{1}$, Ke Wang$^{1}$, Weikang Shi$^{1}$\\ {\bf Junting Pan}$^{1,3}$, {\bf Mingjie Zhan}$^{\dagger1}$, {\bf Hongsheng Li}$^{\dagger1,2,3}$\\
  $^1$Multimedia Laboratory (MMLab), The Chinese University of Hong Kong\\ $^2$Shanghai Artificial Intelligence Laboratory \quad $^3$CPII under InnoHK\\
 \texttt{luzimu@mail.ustc.edu.cn} \quad \texttt{\{aojunzhou, zmjdll\}@gmail.com}  \\ \texttt{hsli@ee.cuhk.edu.hk} 
}

\begin{document}
\maketitle
\begin{abstract}

Large language models (LLMs) have exhibited great potential in mathematical reasoning. However, there remains a performance gap in this area between existing open-source models and closed-source models such as GPT-4. In this paper, we introduce \textbf{MathGenie}, a novel method for generating diverse and reliable math problems from a small-scale problem-solution dataset (denoted as {\it seed data}). We augment the ground-truth solutions of our seed data and train a back-translation model to translate the augmented solutions back into new questions. Subsequently, we generate code-integrated solutions for the new questions. To ensure the correctness of the code-integrated solutions, we employ rationale-based strategy for solution verification. Various pretrained models, ranging from 7B to 70B, are trained on the newly curated data to test the effectiveness of the proposed augmentation technique, resulting in a family of models known as {\it MathGenieLM}. These models consistently outperform previous open-source models across five representative mathematical reasoning datasets, achieving state-of-the-art performance. In particular, MathGenieLM-InternLM2 achieves an accuracy of 87.7\% on GSM8K and 55.7\% on MATH, securing the best overall score among open-source language models.

\end{abstract}
\blfootnote{$^*$Equal contribution\quad $^\dagger$Corresponding author}

\section{Introduction}

\begin{figure*}[tb!]
    \centering
    {\includegraphics[width=.95\textwidth]{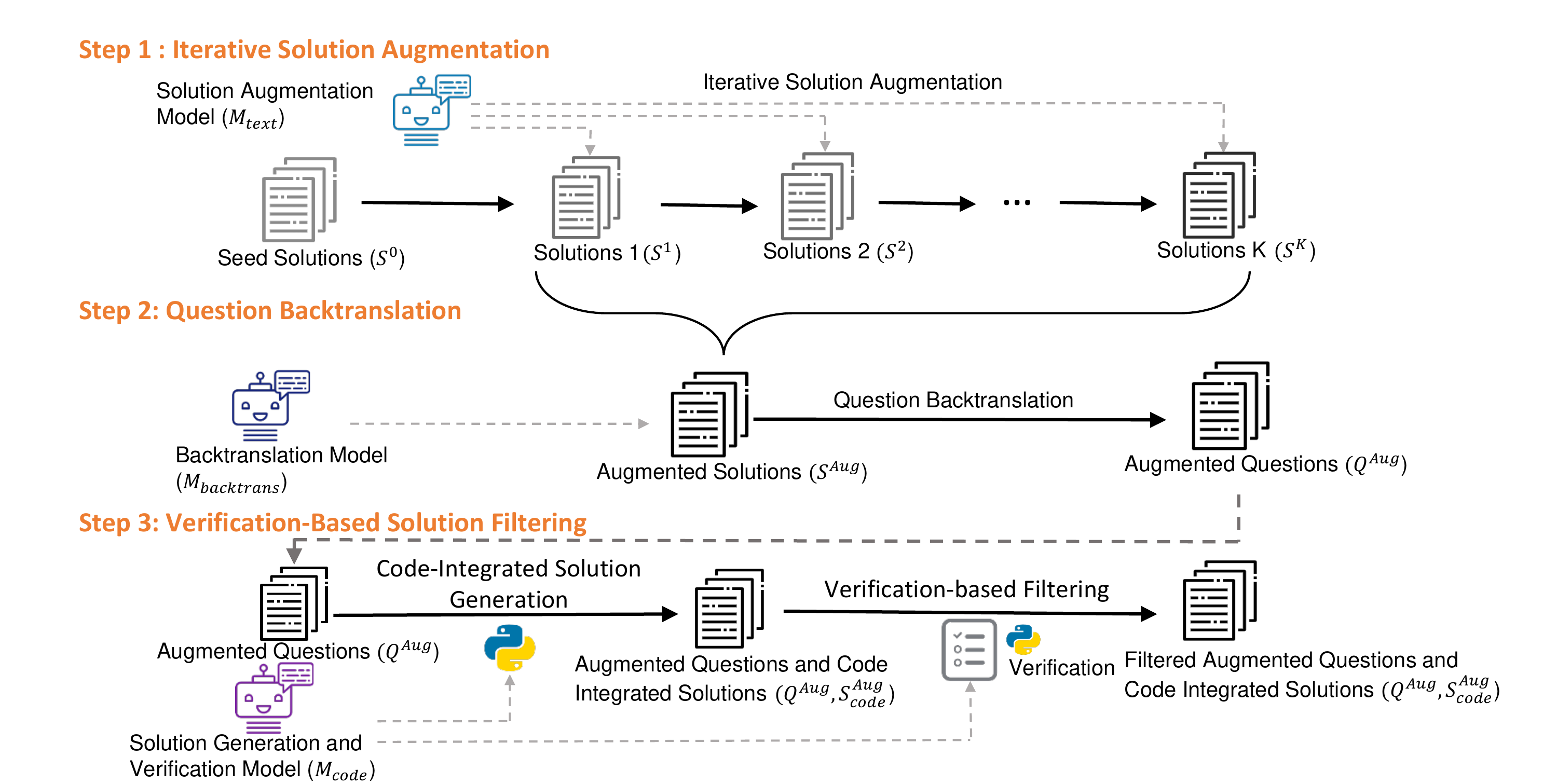}}
    \caption{Framework of MathGenie. Iterative Solution Augmentation augments human-annotated solutions in GSM8K and MATH to create new solutions, as shown in Step 1. These solutions are then back-translated to new questions using Question Back-translation, demonstrated in Step 2. Then reliable code-integrated solutions are curated using Verification-Based Solution Filtering, by generating solutions and filtering them using verification rationales, as shown in Step 3.
    }
    \label{fig:pipeline}
\end{figure*}

Large language models (LLMs), such as GPT-4~\citep{OpenAI2023GPT4TR}, and others~\citep{touvron2023llama,yue2023mammoth,gou2023tora,wang2023mathcoder}, have demonstrated outstanding capabilities in mathematical reasoning. Existing methods explored three main types of solution formats for tackling the math-solving problem: text-only Chain-of-Thought (CoT) solutions~\citep{wei2022chain}, code-only Program-of-Thought (PoT) solutions~\citep{chen2022program}, and code-integrated solutions~\citep{zhou2023solving}. Among these, code-integrated solutions demonstrate superior performance over both CoT and PoT solutions ~\citep{gou2023tora,wang2023mathcoder, zhou2023solving}, indicating their effectiveness in enhancing problem-solving abilities.

In this paper, we focus on creating diverse and reliable augmented questions and ensuring the reliability of the generated code-integrated solutions, thereby better finetuning pretrained base models. Existing works, such as ToRA~\citep{gou2023tora} and MathCoder~\citep{wang2023mathcoder}, build code-integrated solutions and augment math questions using off-the-shelf GPT-4. However, scaling up the training data with GPT-4 becomes prohibitively expensive.

Consequently, developing a free open-source model to generate large-scale synthetic data presents a promising alternative, offering scalability and cost-effectiveness~\citep{yuan2023scaling,singh2023beyond}. To effectively scale-up math problem-solving data, we focus on handling two critical challenges: (1) how to generate high-quality and diverse math problems to aid in generalization, and (2) how to generate accurate and reliable solutions for the augmented problems without human-annotated ground truth, preferably in a code-integrated format.
A unified framework, \textit{MathGenie}, is proposed, which consists of three components as shown in Fig.~\ref{fig:pipeline} to tackle the above-mentioned challenges: \textit{Iterative Solution Augmentation}, \textit{Question Back-translation} and \textit{Verification-Based Solution Filtering}.

Iterative Solution Augmentation and Question Back-translation aims to generate diverse and reliable math problems. Unlike direct question augmentation~\citep{yu2023metamath}, the proposed math problem back-translation leverages the constraints and logical relationships inherent in mathematical solutions to create a diverse and high-quality set of new math problems. Specifically, we iteratively augment the human-annotated solutions from the relatively small training sets of MATH~\cite{math2021} and GSM8K~\cite{gsm8k2021}, generating a large-scale collection of augmented new solutions, as shown in Step 1 of Fig.~\ref{fig:pipeline}. These solutions are then processed by a math back-translation model, $M_{\rm backtrans}$, to back-translate the augmented solutions into their corresponding math questions, as demonstrated in Step 2 of Fig.~\ref{fig:pipeline}. This method draws inspiration from Instruction Back-translation~\citep{li2023self}, which back-translates instructions from texts in a web corpus. However, the key difference is that our source solutions for back-translation are augmented from existing ones to ensure the reliability and solvability of the augmented questions.

\begin{figure*}[t]
    \centering
    \includegraphics[width=0.8\linewidth]{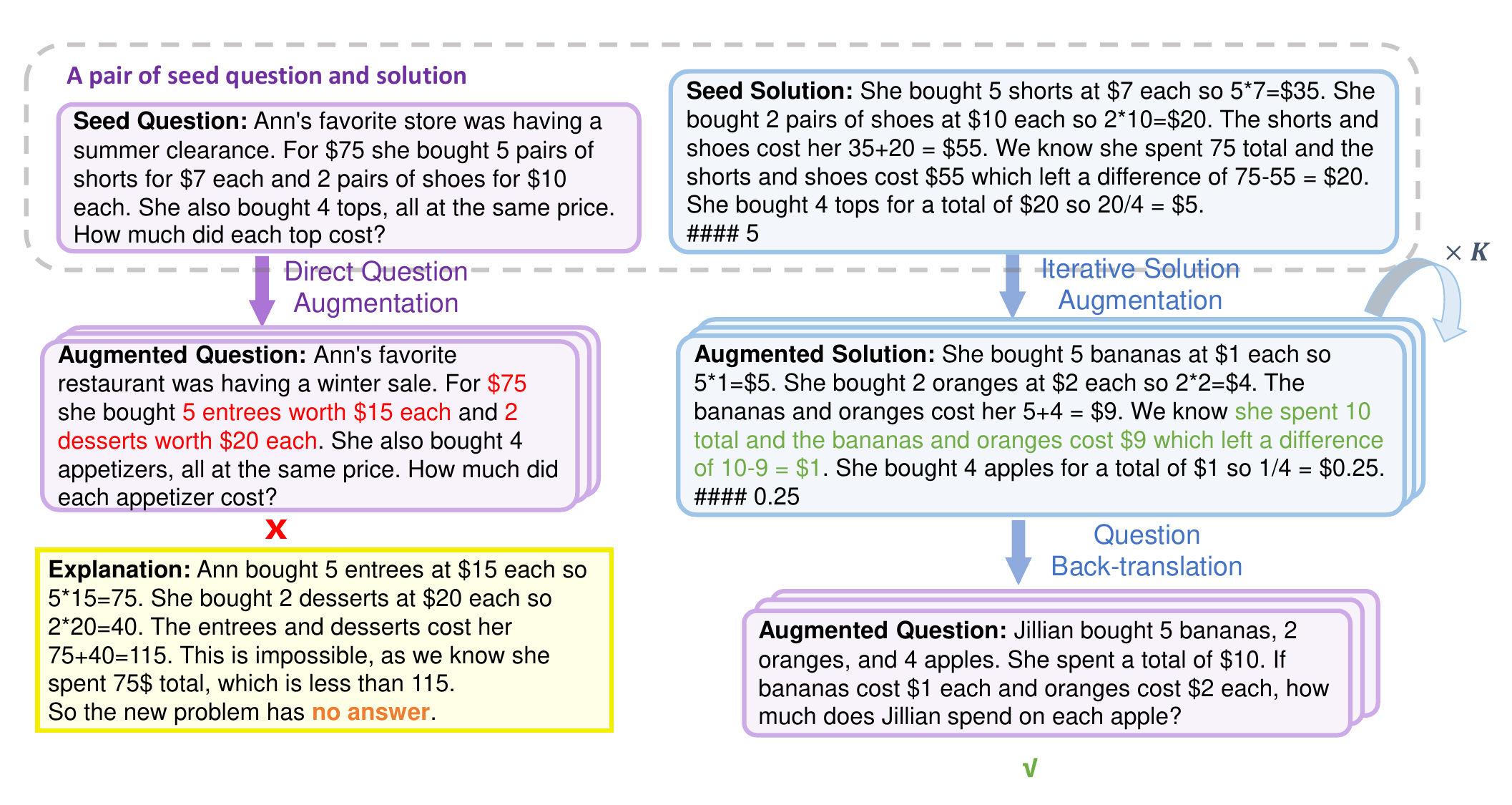}
    \caption{Comparison between Direct Question Augmentation (left) and Iterative Solution Augmentation and Question Back-translation (right). Direct Question Augmentation damages the hidden constraints between the conditions (cost of part of the things bought must not be more than the total cost), thus producing a question with no answer. Question Back-translation considers the solution, and correctly augments the question.}
    
\label{fig:aug_question}
\end{figure*}

These newly generated math problems lack reliable ground-truth solutions, which necessitates the proposed Verification-Based Solution Filtering. We first build a model, $M_{\rm code}$, capable of generating code-integrated solutions and verifying these solutions. Then, code-integrated solutions of the new questions are generated with this model. To enhance the reliability of these code-integrated solutions, we use $M_{\rm code}$ to verify the model-generated solutions by generating verification rationales for them, as demonstrated in Step 3 of Fig.~\ref{fig:pipeline}. The verification rationales use interleaved natural language and code to verify the correctness of the solutions, as shown in Tab.~\ref{tab:example_verify_correct} and Tab.~\ref{tab:example_verify_wrong} in Appendix~\ref{app:verify_examples}. Only the solutions verified to be correct are retained, thus improving the accuracy and quality of the generated data.

Based on the proposed MathGenie framework, we obtain a large-scale model-generated math problem-solution dataset, {\it MathGenieData}, featuring diverse augmented questions and reliable code-integrated solutions.

To evaluate the effectiveness of the question-solution augmentation framework {MathGenie}, we finetune various state-of-the-art pretrained models, ranging from 7B to 70B. This results in \textbf{MathGenieLM}, a new family of math models with excellent performance. Our models demonstrate high accuracy across five diverse and representative math datasets: MATH, GSM8K, SVAMP, Simuleq, and Mathematics. In particular, MathGenieLM-InternLM2 achieves an accuracy of 87.7\% on GSM8K and 55.7\% on MATH, achieving the best overall score. When enhanced by majority voting, MathGenieLM-Llama-2-70B attains a 10-path accuracy rate of 91.7\% on GSM8K and 63.3\% on MATH.

The main contributions of this paper are summarized as follows: (1) We propose the MathGenie pipeline, which is designed to enhance the scale, diversity, and quality of synthetic math questions, as well as to improve the accuracy of the code-integrated solutions generated for them. (2) We conduct extensive experiments on various pretrained language models, demonstrating consistently superior performance across multiple math datasets.

\section{MathGenie}

In this section, we introduce MathGenie, a pipeline for creating diverse and reliable math problems through back-translation, and curating high-quality code-integrated solutions through verification. We begin by introducing the seed data and solution generator model. Next, we present the proposed MathGenie pipeline, which consists of three key steps, as shown in Fig.~\ref{fig:pipeline}: Iterative Solution Augmentation, Question Back-translation, and Verification-Based Solution Filtering.

\textbf{Seed Data.} The seed data consists of two parts: (1) The first part is used for data augmentation, consisting of 15K math problems and human-annotated solutions from the training sets of GSM8K and MATH. We denote it as $\hD_{\rm text}=\{(q^i, s^i)\}_{i=1}^n$, where $q^i$ is the $i$-th question, $s^i$ is its natural-language solution, and $n$ is the total number of cases. (2) The second part is used for training our candidate solution generator model, which serves to generate candidate solutions for the augmented questions. We denote this part of the seed data as $\hD_{\rm code}=\{(q^i, s^i_{\rm code}, v^i_{\rm code})\}_{i=1}^N$, where $q^i$ is the question, $s^i_{\rm code}$ is its code-integrated solution, and $v^i_{\rm code}$ is the code-integrated verification rationales for the question-solution pair. It contains 80K samples of code-integrated solutions for problems in GSM8K and MATH, as well as code-integrated verification rationales for these solutions. Multiple different solutions are collected for each question. We acquire these solutions and verification rationales using the GPT-4 Code Interpreter, which consist of interleaved natural language and code. 

\textbf{Candidate Solution Generator.} The candidate solution generator is a Llama-2 70B model trained with $\hD_{\rm code}$ and denoted as $M_{\rm code}$. The code-integrated solutions in $\hD_{\rm code}$ enables $M_{\rm code}$ to output candidate code-integrated solutions for given math problems, similar to~\cite{wang2023mathcoder}. It has an accuracy of 86.4\% on GSM8K and 49.5\% on MATH. The verification rationales in $\hD_{\rm code}$ enable $M_{\rm code}$ to effectively verify the solutions with code-integrated rationales. The training method of data in the code-integrated format is as described in~\citet{wang2023mathcoder}.

\textbf{Iterative Solution Augmentation.} Different from previous works that directly augment math questions~\cite{luo2023wizardmath}, 
we propose to augment solutions first and then back-translate the augmented solutions to their corresponding questions to better constrain the question generation process and enhance the reliability of machine-generated questions. The proposed strategy is also different from the previous Instruction Back-translation method~\citep{li2023self}, which leverages large amounts of texts in a web corpus. As existing solutions are limited in number and already have corresponding questions, it is necessary to augment them before the back-translation.

To augment the solutions in $\hD_{\rm text}$ into related new ones, we develop a solution augmentation model, $M_{\rm text}$, by finetuning the LLaMA-2 70B model on high-quality instructional datasets, including OpenOrca\footnote{\url{https://huggingface.co/datasets/Open-Orca/OpenOrca}} and Alpaca-GPT4\footnote{\url{https://huggingface.co/datasets/vicgalle/alpaca-gpt4}}. $M_{\rm text}$ takes in a solution and a prompt instructing the model to augment it, and outputs an augmented solution. The prompts are shown in Tab.~\ref{tab:solution_aug_prompts}. The augmentations are carefully constrained and reliable. We iteratively augment each solution in $\hD_{\rm text}$. For convenience, the set of human-annotated solutions in $\hD_{\rm text}$ are denoted as $\hS^0$. The solutions in $\hS^0$ are augmented by $M_{\rm text}$ to create $\hS^1$. As shown in Step 1 of Fig.~\ref{fig:pipeline}, this process is repeated on the previously generated solutions, with $\hS^2$ being created from $\hS^1$, and so on. After $K$ rounds, the final set of augmented solutions, denoted as $\hS^\text{Aug}$, is created by taking the union of $\hS^1, \hS^2, \dots, \hS^K$:
\begin{equation}
    \hS^{\text{Aug}} = \hS^{1}  \cup \hS^{2} \cup \dots \cup \hS^{K}.
\end{equation}

Through iterative solution augmentation, each round produces a set of solutions that differs from the previous, making the solutions gradually deviate more from the original solutions. Consequently, the diversity of the augmented solutions is ensured, which leads to diverse augmented question-solution pairs as mentioned in the following sections. The iteration process is beneficial to the final performance, as demonstrated in Tab.~\ref{tab:iteration_ablation} of Appendix~\ref{app:iteration_ablation}

\begin{table}[t!]\fontsize{7.3}{2}\selectfont
\centering
\begin{tabularx}{\columnwidth}{X}
\toprule 
\textbf{Prompt} \\
\midrule
\textit{Replace the objects and verbs} in Solution. \\
Solution:\\\\\{solution\} \\\\
You must replace the objects and verbs in the Solution with objects and verbs very different from before to create a new solution.\\
\midrule
\textit{Add an extra step} to the solution so that it is more complicated. \\
Solution:\\\\\{solution\} \\\\
You must add an extra step to the Solution to create a new solution. \\
\midrule
\textit{Change the numbers} in Solution. \\
Solution:\\\\\{solution\} \\\\
You must change the numbers in the Solution to create a new solution.\\
\midrule
\textit{Replace the numbers and variables} in Solution with different variables and numbers you pick randomly. \\
Solution:\\\\\{solution\} \\\\
You must replace the numbers and variables in the Solution with different numbers and variables you pick randomly to create a new solution.\\
\bottomrule 
\end{tabularx}
\caption{Different prompts for solution augmentation. ``\{solution\}'' is replaced with the solution to be augmented.}
\label{tab:solution_aug_prompts}
\end{table}

\textbf{Question Back-translation.} We introduce Question Back-translation to translate the solutions in $\hS^{\text{Aug}}$ back to their corresponding math problems. To enhance the accuracy of the translation, we build a Question Back-translation Model $M_{\rm backtrans}$ by finetuning Llama-2 70B on the reversed pairs of question and solution in $\hD_{\rm text}$. The format of each sample in the finetuning data can be denoted as $(s, q)$, where $s$ is a solution, serving as input, and $q$ is its corresponding question. Then, we present the solutions in $\hS^{\text{Aug}}$ to $M_{\rm backtrans}$, thereby translating them back to a set of new questions, denoted as $\hQ^{\text{Aug}}$. An example of Question Back-translation is presented in Tab.~\ref{tab:backtranslation_example}.

\begin{table}[t]\fontsize{7.3}{2}\selectfont
\centering
\begin{tabularx}{\columnwidth}{X}
\toprule 
\textbf{Input} \\
\midrule
Create a New Problem based on the Solution:\\
\\
For the soccer practice, Jack needs 3 days * 10 soccer balls/day = 30 soccer balls.
Adding the soccer balls for the Gym class, he needs to give a total of 30 soccer balls + 10 soccer balls = 40 soccer balls.\\
\midrule
\textbf{Output} \\
\midrule
New Problem: Jack is in charge of the equipment room at his school. He needs to gather 10 soccer balls for the gym class and also give 3 days worth of soccer balls for the soccer team’s practice. If each practice uses 10 soccer balls, how many soccer balls does Jack need to give out?\\
\bottomrule 
\end{tabularx}
\caption{An example of the proposed Question Back-translation. The solution is input to the question back-translation model, and the model outputs its corresponding question.}
\label{tab:backtranslation_example}
\end{table}

Question Back-translation operates on $\hS^{\text{Aug}}$, which is less unpredictable than the texts from the web corpus used in Instruction Back-translation~\cite{li2023self}. By leveraging the constraints shown in the solutions, it is possible to create new questions more reliable than what direct question augmentation can produce, as validated in experiments.

\textbf{Verification-based Solution Filtering.} Existing open-source models such as MathCoder~\citep{wang2023mathcoder} only has the ability to solve math problems but are unable to effectively verify their solutions. 
We enhance $M_{\rm code}$'s ability to verify the solutions by adding code-integrated verification rationales to the finetuning data. The training samples of verification rationales in the seed data are in the format of $(q, s_{\rm code}, v_{\rm code})$, where $q$ and $s_{\rm code}$ are a pair of question and code-integrated solution, and $v_{\rm code}$ is the code-integrated verification. The ($q, s_{\rm code}$) pairs are the input, while the model is trained to output $v_{\rm code}$. In this way, the baseline math problem-solving model $M_{\rm code}$ acquires the ability to verify its solutions with rationales made of interleaved natural language and code. This ability not only facilitates Verification-Based Solution Filtering, but can also play a role in enhancing inference accuracy.

To perform the proposed Verification-Based Solution Filtering, we first generate code-integrated solutions for each question in $\hQ^\text{Aug}$. Initial filtering is performed using answer consistency~\cite{wang2022self}, removing a question if its solutions reach different answers. We then present each question-solution pair to $M_{\rm code}$, prompting it to output a code-integrated verification rationale, from which we can determine whether the solution is verified as correct or wrong. Examples of the verification process are shown in Appendix~\ref{app:verify_examples}. Candidate solutions that are verified to be wrong are abandoned. The process is demonstrated in Step 3 of Fig.~\ref{fig:pipeline}.

The pipeline proposed above results in 170K samples of question and code-integrated solution pairs, denoted as AugData. AugData consists of two parts: the 110K samples augmented from GSM8K dataset, denoted as AugGSM8K, and the 60K samples augmented from MATH dataset, denoted as AugMATH. We denote the above mentioned seed data for training $M_{\rm code}$ as SeedData. Combining SeedData and AugData, we present the final dataset, \textbf{MathGenieData}, which can be used to finetune various pretrained models, such as Llama-2~\citep{touvron2023llama} and CodeLlama~\citep{roziere2023code}, enhancing their problem-solving ability and solution verification skills. The resulting family of mathematical reasoning models is named MathGenieLM.

\section{Experiments}

\begin{table*}[t!]
\fontsize{8.5}{9}\selectfont
\centering
\begin{tabularx}{\textwidth}{@{}*{2}{>{\centering\arraybackslash\hsize=1.2\hsize}X} > {\centering\arraybackslash\hsize=.4\hsize}X |*{2}{>{\centering\arraybackslash\hsize=\hsize}X} | *{2}{>{\centering\arraybackslash\hsize=\hsize}X} > {\centering\arraybackslash\hsize=1.2\hsize}X | >{\centering\arraybackslash\hsize=\hsize}X@{}}
\midrule
\multirow{2}{*}{\textbf{Model}}      & \multirow{2}{*}{\textbf{Base}}         & \multirow{2}{*}{\textbf{Size}}    & \multicolumn{2}{c|}{\textbf{In-Domain}}  & \multicolumn{3}{c|}{\textbf{Out-of-Domain}}   &   \\

&      &   & \textbf{GSM8K}   & \textbf{MATH}   & \textbf{SVAMP}   & \textbf{Simuleq}        & \textbf{Mathematics}   & \textbf{Average}   \\
\midrule
\multicolumn{9}{c}{\textbf{Colsed-Source Models}}    \\
\midrule
ChatGPT-3.5      & -         & -    & 80.8   & 35.5   &  83.0   & -        & -   & -  \\
GPT-4      & -         & -    & 92.0    & 42.5   & 97.0    & -        & -   & - \\
GPT-4 Code      & -         & -    & 97.0    & 69.7   & -   & -        & -   & - \\
PaLM-2       & -         & -    & 80.7     & 34.3    & -   & -        & -   & - \\
\midrule
\multicolumn{9}{c}{\textbf{Open-Source Models}}    \\
\midrule
Mammoth      & CodeLlama         & 7B    & 59.4   & 33.4   &  71.4   & 45.9        & 55.4   & 53.1  \\
MathCoder      & CodeLlama         & 7B    & 67.8   & 30.2    &  70.7   & 49.6       & 55.8   & 54.8  \\
ToRA      & CodeLlama         & 7B    & 72.6   & 44.6   &  70.4   & 36.0        & 68.1   & 58.3  \\
\rowcolor{blue!5}
\textbf{MathGenieLM}       & CodeLlama         & 7B    & 71.5   & 39.7   &  80.2   & 69.1        & 69.5   & 
66.0  \\
Mammoth      & Llama-2         & 7B    & 53.6   & 31.5   &  67.7   & 41.2        & 46.3   & 48.1  \\
MathCoder      & Llama-2         & 7B    & 64.2   & 23.3    &  71.5   & 47.5       & 46.9   & 50.7  \\
ToRA      & Llama-2         & 7B    & 68.8   & 40.1   &  68.2   & 29.2        & 58.3   & 52.9  \\
\rowcolor{blue!5}
\textbf{MathGenieLM}      & Llama-2         & 7B    & 71.7   & 33.0   &  78.5   & 61.4        & 65.0   & 
61.9  \\
\rowcolor{blue!5}
\textbf{MathGenieLM}      & Llemma         & 7B    & \underline{76.0}   & \textbf{48.3}   &  \underline{81.3}   & \textbf{85.0}       & \textbf{76.6}   & \textbf{73.4}  \\
\rowcolor{blue!5}
\textbf{MathGenieLM}      & Mistral         & 7B    & \textbf{80.5}   & \underline{45.1}   &  \textbf{83.3}   & \underline{79.4}       & \underline{71.8}   & \underline{72.0}  \\
\midrule
Mammoth      & CodeLlama         & 13B    & 64.7   & 36.3   &  73.7   & 47.1        & 61.5   & 56.7  \\
MathCoder      & CodeLlama         & 13B    & 74.1   & 35.9    &  78.0   & 60.7       & 62.5   & 62.2  \\
ToRA      & CodeLlama         & 13B    & 75.8   & \textbf{48.1}   &  75.7   & 37.9        & 70.3   & 61.6  \\
\rowcolor{blue!5}
\textbf{MathGenieLM}     & CodeLlama         & 13B    & \underline{78.5}   & 40.3   &  \textbf{84.5}   & \underline{76.7}        & \underline{65.7}   & 
\underline{69.1}  \\
Mammoth      & Llama-2         & 13B    & 62.0   &  34.2   &  72.4   & 43.2        & 49.2   & 52.2  \\
MathCoder      & Llama-2         & 13B    & 72.6   & 29.9    &  76.9   & 62.3       & 54.7   & 59.3  \\
ToRA      & Llama-2         & 13B    & 72.7   & 43.0   &  72.9   & 45.7        & 62.6   & 59.4  \\
\rowcolor{blue!5}
\textbf{MathGenieLM}      & Llama-2         & 13B    & \textbf{80.4}   & \underline{43.8}   &  \underline{83.5}   & \textbf{78.4}        & \textbf{72.7}   & 
\textbf{71.8}  \\
\midrule
Mammoth      & CodeLlama         & 34B    & 72.7   & 43.6   &  84.3   & 51.8        & 65.4   & 63.6  \\
MathCoder      & CodeLlama         & 34B    & 81.7   & 45.2    &  82.5   & 65.8       & 75.9   & 70.2  \\
ToRA      & CodeLlama         & 34B    & 80.7   & 50.8   &  80.5   & 50.2        & 77.9   & 68.0  \\
\rowcolor{blue!5}
\textbf{MathGenieLM}       & CodeLlama         & 34B    & \textbf{86.2}   & \underline{51.4}   &  \underline{86.9}   & \underline{85.8}        & \underline{78.4}   & \underline{77.7}  \\
\rowcolor{blue!5}
\textbf{MathGenieLM}      & Llemma         & 34B    & \underline{84.1}   & \textbf{55.1}   &  \textbf{87.4}   & \textbf{89.3}       & \textbf{82.9}   & \textbf{79.8}  \\
\midrule
WizardMath      & Llama-2         & 70B    & 81.6   & 22.7   &  -   & -       & -   & -  \\
Mammoth      & Llama-2         & 70B    & 76.9   & 41.8   &  82.4   & 51.4       & 55.6   & 61.6  \\
MathCoder      & Llama-2         & 70B    & 83.9    & 45.1   &  \underline{84.9}    & \underline{77.0}       & \underline{74.4}    & \underline{73.1}  \\
ToRA      & Llama-2         & 70B    & \underline{84.3}   & \underline{49.7}   &  82.7   & 73.3       & 72.6   & 72.5  \\
\rowcolor{blue!5}
\textbf{MathGenieLM}      & Llama-2         & 70B    & \textbf{88.4}   & \textbf{51.2}   &  \textbf{87.7}   & \textbf{89.1}        & \textbf{76.0}   & \textbf{78.5}  \\
\midrule
\rowcolor{blue!5}
\textbf{MathGenieLM}      & Mixtral         & 8x7B    & 87.0   & 53.7   &  88.9   & 89.9        & 81.7   & 80.2  \\
\rowcolor{blue!5}
\textbf{MathGenieLM}   & InternLM2     & 20B    & 87.7   & 55.7   &  87.3   & 88.5      & 85.1   & 80.9  \\
\midrule
\end{tabularx}
\caption{Results of MathGenieLM, compared to various open-source and closed-source models on 2 in-domain datasets (GSM8K, MATH), and 3 out-of-domain datasets (SVAMP, Simuleq, Mathematics). The results of closed-source models are from~\citet{yue2023mammoth} and ~\citet{gou2023tora}. }
\label{tab:model_performance}
\end{table*}

\begin{table*}[t!]\fontsize{8.5}{9}\selectfont
\centering
\begin{tabularx}{\textwidth}{@{}*{2}{>{\centering\arraybackslash\hsize=1.2\hsize}X} >{\centering\arraybackslash\hsize=.4\hsize}X >{\centering\arraybackslash\hsize=.6\hsize}X |*{2}{>{\centering\arraybackslash\hsize=.85\hsize}X} | *{2}{>{\centering\arraybackslash\hsize=.85\hsize}X}  >{\centering\arraybackslash}X | >{\centering\arraybackslash\hsize=.7\hsize}X@{}}
\midrule
\multirow{2}{*}{\textbf{Model}} & \multirow{2}{*}{\textbf{Base}}  & \multirow{2}{*}{\textbf{Voting}} & \multirow{2}{*}{$k$-\textbf{path}} & \multicolumn{2}{c|}{\textbf{In-Domain}}   & \multicolumn{3}{c|}{\textbf{Out-of-Domain}} &  \\
&   &  &  & \textbf{GSM8K} &   \textbf{MATH}   & \textbf{SVAMP}   & \textbf{Simuleq}  & \textbf{Mathematics} & \textbf{Average} \\
\midrule
ToRA & \mbox{Llama-2 70B}  & \cmark & 50 & 88.3 & 56.9 & - & - & - & - \\
\midrule
\textbf{MathGenieLM} & \mbox{Llama-2 70B}  & \xmark & - & 88.4 & 51.2 & 87.7 & 89.1 & 76.0 & 78.5 \\
\midrule
\textbf{MathGenieLM} & \mbox{Llama-2 70B}  & \cmark & 10 & 91.7\tiny\color{red}({+3.3}) & 63.3\tiny\color{red}({+12.1}) & 94.0\tiny\color{red}({+6.3}) & 95.9\tiny\color{red}({+6.8}) & 87.2\tiny\color{red}({+11.2}) & 86.4\tiny\color{red}({+7.9})\\
\midrule
\end{tabularx}
\caption{Results of naive majority voting. $k$-path  represents the number of solutions generated for majority voting.}
\label{tab:voting}
\end{table*}

\subsection{Experimental Setup}


\textbf{Datasets.}
We evaluate our models on two in-domain datasets: GSM8K~\citep{gsm8k2021} and MATH~\citep{math2021}, whose training sets are used for finetuning. Additionally, we evaluate the final models on three out-of-domain datasets: SVAMP~\citep{patel2021nlp}, Simuleq~\citep{KoncelKedziorski2016MAWPSAM}, and Mathematics~\citep{Davies2021AdvancingMB}, to evaluate the generalization ability of our proposed method.

\textbf{Models.}
We perform full-parameter finetuning on various pretrained models, including Llama-2 7B, 13B, and 70B~\citep{touvron2023llama}, CodeLlama 7B, 13B, and 34B~\citep{roziere2023code}, Llemma 7B and 34B~\citep{azerbayev2023llemma}, Mistral 7B~\citep{jiang2023mistral}, Mixtral-8x7B~\citep{jiang2024mixtral}, and InternLM2 20B~\citep{2023internlm}. Finetuning details are described in Appendix~\ref{app:finetune_detail}.

\textbf{Compared methods.}
We compare MathGenieLM with closed-source models such as ChatGPT-3.5~\citep{brown2020language}, GPT-4~\citep{OpenAI2023GPT4TR}, and PaLM-2~\citep{anil2023palm}, as well as open-source models such as Mammoth~\citep{yue2023mammoth}, MathCoder~\citep{wang2023mathcoder}, ToRA~\citep{gou2023tora}, and WizardMath~\citep{luo2023wizardmath}.

\begin{table*}[t!]\fontsize{7.3}{9}\selectfont
\centering
\begin{tabularx}{\textwidth}{@{}*{4}{>{\centering\arraybackslash}X}|*{2}{>{\centering\arraybackslash}X}|*{3}{>{\centering\arraybackslash}X}|*{1}{>{\centering\arraybackslash}X}@{}}
\toprule 
\multicolumn{4}{c|}{\textbf{Data}} & \multicolumn{2}{c|}{\textbf{In-Domain}} & \multicolumn{3}{c|}{\textbf{Out-of-Domain}}\\
\textbf{Seed} & \textbf{Verification} & \textbf{AugGSM8K} & \textbf{AugMATH} & \textbf{GSM8K} & \textbf{MATH} & \textbf{SVAMP} & \textbf{Simuleq} & \textbf{Mathematics} & \textbf{Average} \\
\midrule
\cmark & \xmark & \xmark & \xmark & \cellcolor{lightest}73.5 & \cellcolor{lightest}41.8 & \cellcolor{lightest}73.1 & \cellcolor{lightest}68.5 & \cellcolor{light}66.6 & \cellcolor{lightest}64.7 \\
\cmark & \cmark & \xmark & \xmark & \cellcolor{light}74.2 & \cellcolor{light}42.2 & \cellcolor{light}76.5 & \cellcolor{light}71.0 & \cellcolor{lightest}65.5 & \cellcolor{light}65.9 \\
\cmark & \cmark & \cmark & \xmark & \cellcolor{dark}78.2 & \cellcolor{light}42.2 & \cellcolor{darkest}84.3 & \cellcolor{dark}78.8 & \cellcolor{medium}69.1 & \cellcolor{dark}70.5 \\
\cmark & \cmark & \xmark & \cmark & \cellcolor{medium}74.9 & \cellcolor{dark}43.5 & \cellcolor{medium}81.6 & \cellcolor{medium}73.3 & \cellcolor{darkest}73.2 & \cellcolor{medium}69.3 \\
\cmark & \cmark & \cmark & \cmark & \cellcolor{darkest}80.5 & \cellcolor{darkest}45.1 & \cellcolor{dark}83.3 & \cellcolor{darkest}79.4 & \cellcolor{dark}71.8 & \cellcolor{darkest}72.0 \\
\midrule\midrule
\multicolumn{4}{l|}{\textbf{Seed + Verification}}   & \cellcolor{lightest}74.2  &   \cellcolor{light}42.2   & \cellcolor{lightest}76.5   & \cellcolor{lightest}71.0    & \cellcolor{lightest}65.5   & \cellcolor{lightest}65.9   \\
\multicolumn{4}{l|}{\textbf{Seed + Verification + $\frac{1}{8}$(AugGSM8K + AugMATH)}}   & \cellcolor{light}75.6  &  \cellcolor{lightest}41.6   & \cellcolor{light}80.7   & \cellcolor{light}75.3    & \cellcolor{light}67.7   & \cellcolor{light}68.2   \\
\multicolumn{4}{l|}{\textbf{Seed + Verification + $\frac{1}{4}$(AugGSM8K + AugMATH)}}   & \cellcolor{medium}77.9 &   \cellcolor{medium}42.7   & \cellcolor{dark}82.6   & \cellcolor{medium}77.0        & \cellcolor{medium}67.8   & \cellcolor{medium}69.6   \\
\multicolumn{4}{l|}{\textbf{Seed + Verification + $\frac{1}{2}$(AugGSM8K + AugMATH)}}   & \cellcolor{dark}79.2 &   \cellcolor{dark}44.0   & \cellcolor{medium}82.1   & \cellcolor{dark}80.2        & \cellcolor{dark}68.8   & \cellcolor{dark}70.9   \\
\multicolumn{4}{l|}{\textbf{Seed + Verification + (AugGSM8K + AugMATH)}}   & \cellcolor{darkest}80.5  &   \cellcolor{darkest}45.1   & \cellcolor{darkest}83.3   & \cellcolor{darkest}79.4   & \cellcolor{darkest}71.8   & \cellcolor{darkest}72.0   \\
\bottomrule 
\end{tabularx}
\caption{Effect of different data composition and amounts of augmented data with Mistral-7B as the base model.}
\label{tab:different_composition}
\end{table*}

\subsection{Main Results}

Tab.~\ref{tab:model_performance} shows the accuracy of MathGenieLM across five datasets. Based on the results, we make the following observations: (1) For open-source models with parameters ranging from 7B to 70B, MathGenieLM achieves state-of-the-art performance. (2) MathGenieLM demonstrates particularly high performance on the three out-of-domain datasets compared to previous open-source models, showcasing the superior generalization capability of our method. (3) MathGenieLM's accuracy exceeds that of ChatGPT-3.5 and PaLM-2. However, there remains a noticeable gap when compared to GPT-4's performance. (4) MathGenieLM-Llemma-34B and MathGenieLM-InternLM2-20B reach over 55\% accuracy on the challenging MATH dataset. This might be attributed to the high-quality math-related data they used in pretraining. (5) Mixtral-8x7B achieves excellent performance, demonstrating the potential of Mixture of Experts (MoE) models. The results in Tab.~\ref{tab:model_performance} are all obtained using greedy decoding.

Apart from the results obtained with greedy decoding, we also report the results of majority voting using multiple sampled paths~\citep{wang2022self}, conducted on MathGenieLM-Llama-2-70B, compared with ToRA-Llama-2-70B. The results are shown in Tab.~\ref{tab:voting}, where ``$k$'' represents the number of solutions generated for majority voting. We observe that, with $k=10$, majority voting significantly increases the accuracy across all five datasets, yielding an average gain of 7.9\%. Specifically, at $k=10$, MathGenieLM-Llama-2-70B achieves an accuracy of 91.5\% on GSM8K and 63.3\% on MATH, significantly outperforming ToRA-70B at $k=50$. This demonstrates the superior performance of our model.

\subsection{Ablation Study}

The following are some ablation studies. All finetuning in the ablation studies has been conducted using Mistral-7B as the base model.

\textbf{Analysis of different data composition.} We analyze the effect of adding and subtracting different parts of our training data to observe the impact of each component. As shown in the upper half of Tab.~\ref{tab:different_composition}, when only AugGSM8K is added, the performance on GSM8K, SVAMP, and Simuleq improves, while adding AugMATH leads to more notable improvements in MATH and Mathematics. This is consistent with the types of questions in each dataset: GSM8K, SVAMP, and Simuleq contain grade-school level math word problems with relatively easy calculations, whereas MATH and Mathematics feature more complex math computations. When both AugGSM8K and AugMATH are added, the improvements in the datasets are compounded as well, which shows the effectiveness of our augmented data.

\begin{figure}[t]
    \centering
    \includegraphics[width=\columnwidth]{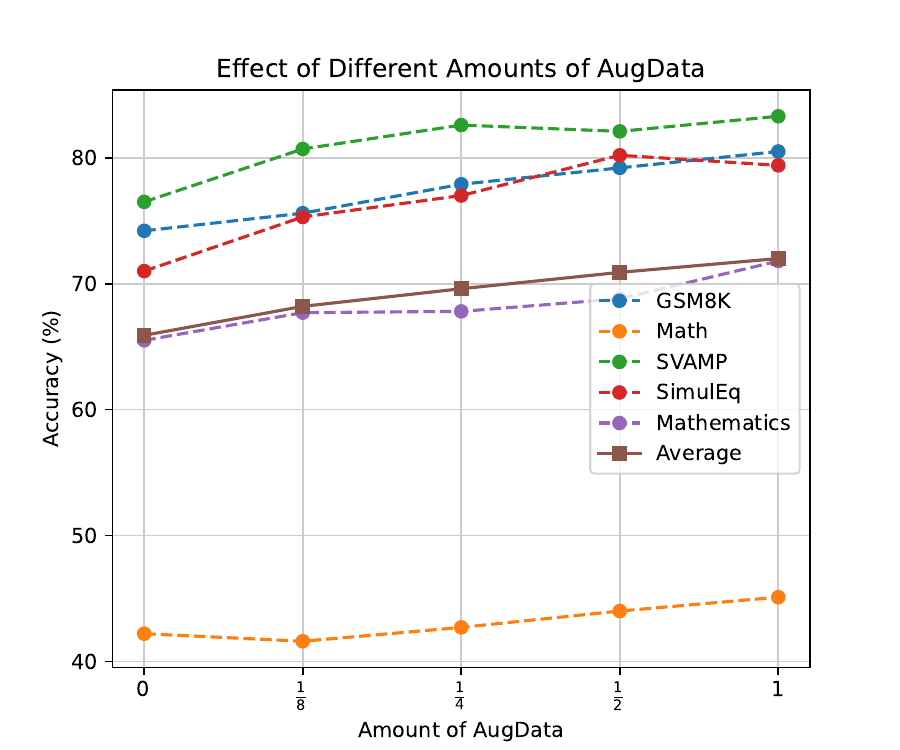}
    \caption{Performance of the Mistral 7B model finetuned with $\{0, \frac18, \frac14, \frac12, 1\}$ times the amount of augmented data.}
    
\label{fig:different_amount}
\end{figure}

\textbf{Analysis of different amounts of augmented data.} We analyze the scaling quality of the augmented data we generated by training a model with \{$0, \frac{1}{8}, \frac{1}{4}, \frac{1}{2}, 1$\} times the amount of augmented data. The results, as shown in the bottom half of Tab.~\ref{tab:different_composition} and Fig.~\ref{fig:different_amount}, indicate that, with an increase in the amount of augmented data, the performance on all five datasets consistently improves, with very few exceptions. This demonstrates the high scaling quality of our data.

\textbf{Analysis of Verification-Based Solution Filtering.} We analyze the effectiveness of Verification-Based Solution Filtering by using data before and after filtering with the help of verification to finetune the model. As demonstrated in Tab.~\ref{tab:verify_filter_effect}, finetuning the model with augmented question-solution pairs filtered by verification results in noticeable accuracy increases in both GSM8K and MATH, indicating the superior quality of the augmented data after filtering and the effectiveness of Verification-Based Solution Filtering. Further analysis of the verification ability of $M_{\text{code}}$ is shown in Tab.~\ref{tab:verify_precision} of Appendix~\ref{sec:code_integrated_verification}.

\begin{table}[t]\fontsize{8.5}{9}\selectfont
\centering
\begin{tabularx}{\columnwidth}{@{}*{2}{>{\centering\arraybackslash}X}|*{2}{>{\centering\arraybackslash}X}|*{1}{>{\centering\arraybackslash}X}@{}}
\midrule
\multicolumn{2}{l|}{\textbf{Data}}   & \textbf{GSM8K} &   \textbf{MATH}  & \textbf{Average}   \\
\midrule
\multicolumn{2}{l|}{\textbf{w/o verification filtering}}   & 79.3  &  43.8   & 61.6   \\
\midrule
\multicolumn{2}{l|}{\textbf{w/ verificatioin filtering}}   & 80.5\tiny\color{red}({+1.2})  &   45.1\tiny\color{red}({+1.3})     & 62.8\tiny\color{red}({+1.2})  \\
\midrule
\end{tabularx}
\caption{Effect of using or not using code-integrated verification rationales to filter the training data. }
\label{tab:verify_filter_effect}
\end{table}

\begin{table*}[t]\fontsize{8.5}{9}\selectfont
\centering
\begin{tabularx}{\textwidth}{@{}*{2}{>{\centering\arraybackslash}X}|*{2}{>{\centering\arraybackslash\hsize=.85\hsize}X} | *{2}{>{\centering\arraybackslash\hsize=.85\hsize}X}  >{\centering\arraybackslash}X|*{1}{>{\centering\arraybackslash}X}@{}}
\midrule
\multicolumn{2}{c|}{\multirow{2}{*}{\textbf{Augmentation Method}}} & \multicolumn{2}{c|}{\textbf{In-Domain}} & \multicolumn{3}{c|}{\textbf{Out-of-Domain}}\\
& & \textbf{GSM8K} &   \textbf{MATH}   & \textbf{SVAMP}   & \textbf{Simuleq}        & \textbf{Mathematics}   & \textbf{Average}   \\
\midrule
\multicolumn{2}{l|}{\textbf{MetaMath}}   & 79.5  &  44.6   & 78.4   & 79.6   & 67.9   & 70.0  \\
\midrule
\multicolumn{2}{l|}{\textbf{Direct question augmentation (w/o sol.)}}   & 78.8  &   43.0   & 84.0   & 77.0   & 70.2   & 70.6  \\
\midrule
\multicolumn{2}{l|}{\textbf{Direct question augmentation (w/ sol.)}}   & 79.2  &   44.2   &  83.6  &  72.0  &  68.6  & 69.5  \\
\midrule
\multicolumn{2}{l|}{\textbf{MathGenie (ours)}}   & \textbf{80.5}  &  \textbf{45.1}   & \textbf{83.3}   & \textbf{79.4}    & \textbf{71.8}   & \textbf{72.0}   \\
\midrule
\end{tabularx}
\caption{Comparison of different question augmentation methods. \textbf{Direct question augmentation (w/o sol.)} presents $M_{\rm text}$ with only the seed questions to generate new questions. \textbf{Direct question augmentation (w/ sol.)} presents $M_{\rm text}$ with pairs of question-solution to generate new questions.}
\label{tab:question_aug_methods}
\end{table*}

\textbf{Comparison with other question augmentation methods.}
We compare our method with three other question augmentation methods: MetaMath~\citep{yu2023metamath}, direct question augmentation without solution, and direct question augmentation with solution. The two direct question augmentation methods both utilize $M_{\rm text}$ as the question augmentation model. The former presents only the seed question to the model during question augmentation, while the latter presents both the question and its solution. The results, as shown in Tab.~\ref{tab:question_aug_methods}, indicate that our augmented solution-to-question back-translation method yields better performance than existing augmentation methods.

\subsection{Accuracy of Verified Inference}
\label{sec:verified_inference}

\begin{table*}[t!]\fontsize{8.5}{9}\selectfont
\centering
\begin{tabularx}{\textwidth}{@{}*{2}{>{\centering\arraybackslash}X}|*{2}{>{\centering\arraybackslash}X}|*{3}{>{\centering\arraybackslash}X}|*{1}{>{\centering\arraybackslash}X}@{}}
\midrule
&   & \multicolumn{2}{c|}{\textbf{In-Domain}}   & \multicolumn{3}{c|}{\textbf{Out-of-Domain}} &  \\
&  & \textbf{GSM8K} &   \textbf{MATH}   & \textbf{SVAMP}   & \textbf{Simuleq}        & \textbf{Mathematics}   & \textbf{Average}   \\
\midrule
\multirow{2}{*}{\textbf{Baseline}}  & \textbf{Accuracy}   & 88.4   & 51.2   &  87.7   & 89.1        & 76.0   & 78.5  \\
& \textbf{N$\times$}   & \textcolor{gray}{1$\times$}  &   \textcolor{gray}{1$\times$}   & \textcolor{gray}{1$\times$}   & \textcolor{gray}{1$\times$}   & \textcolor{gray}{1$\times$}   & \textcolor{gray}{1$\times$} \\
\midrule
\multirow{2}{*}{\textbf{Verify (twice)}}  & \textbf{Accuracy}   & 88.6  &  55.8   & 88.7   & 91.2    & 81.1   & 81.1  \\
& \textbf{N$\times$}   & \textcolor{gray}{2.1$\times$}  &   \textcolor{gray}{2.8$\times$}   & \textcolor{gray}{2.1$\times$}   & \textcolor{gray}{2.1$\times$}   & \textcolor{gray}{2.3$\times$}   & \textcolor{gray}{2.3$\times$}  \\
\midrule
\multirow{2}{*}{\textbf{Voting (3-path)}}  & \textbf{Accuracy}   & 88.6  &  53.8   & 92.0   & 90.9    & 81.5   & 81.4   \\
& \textbf{N$\times$}   & \textcolor{gray}{3$\times$}  &   \textcolor{gray}{3$\times$}   & \textcolor{gray}{3$\times$}   & \textcolor{gray}{3$\times$}   & \textcolor{gray}{3$\times$}   & \textcolor{gray}{3$\times$}  \\
\midrule
\end{tabularx}
\caption{Result of MathGenieLM-Llama-2-70B using verified inference. \textbf{Verify (twice)} means that, when testing the model, the solutions are verified and those verified as incorrect are solved again. This process is repeated twice. Each verification or solution is counted as 1 generation, and N$\times$ is the average generation count of each question.}
\label{tab:verified_inference}
\end{table*}

Our models have the ability to verify its own solutions when presented with prompts as shown in Tab.~\ref{tab:verification_format}. This represents a mathematical reasoning ability that can be applied during inference.

A simple way to do this is to verify the solutions generated and solve the problem again if the solution is verified to be incorrect. We limit the number of verification to two times. 
As shown in Tab.~\ref{tab:verified_inference}, applying verification twice consistently enhances accuracy across all five datasets, with notable improvements in the MATH and Mathematics datasets. The average generation (N$\times$) presented in Tab.~\ref{tab:verified_inference} measures the cost of verified inference, which is 2.3$\times$ on average. When compared to 3-path majority voting, verified inference achieves almost identical accuracy but at a significantly reduced cost. The results of more rounds of verification are analyzed in Tab.~\ref{tab:different_verified_inference_rounds} of Appendix~\ref{app:verification_rounds}.

\section{Related Works}

\textbf{Large Language Models for Mathematical Reasoning.}
LLMs have demonstrated remarkable performance in mathematical reasoning tasks. CoT~\citep{wei2022chain} enhances LLMs' capability for multistep reasoning. Self-Consistency~\citep{wang2022self} selects the final answer through majority voting. CSV~\citep{zhou2023solving} introduces code-based self-verification. Other research efforts focus on pretraining or finetuning LLMs, thereby producing math-specific LLMs, such as Llemma~\citep{azerbayev2023llemma}, WizardMath~\citep{luo2023wizardmath}, Mammoth~\citep{yue2023mammoth}, ToRA~\citep{gou2023tora}, and MathCoder~\citep{wang2023mathcoder}. It is noted that code-integrated models~\citep{wang2023mathcoder,gou2023tora} have shown superior capabilities over CoT-style models. This paper develops synthetic math problems and solutions using free models to enhance mathematical reasoning.

\noindent
\textbf{Instruction-following Datasets for LLMs.}
Recent studies~\citep{alpaca,peng2023instruction,mukherjee2023orca,li2023self} have begun to utilize synthetic instructions generated by LLMs, such as GPT-4 or GPT-3.5, to distill into smaller models. WizardLM~\citep{xu2023wizardlm} proposes complex instructions to enrich the seed data for general chat models. This paper, however, focuses on math problem augmentation, particularly for code-integrated math-specific models.

\noindent
\textbf{Data Augmentation for Mathematical Reasoning.}
To upscale the number of math problems, various works~\citep{yu2023metamath,liu2024augmenting,li2023query} directly augment existing problems. Differing from these approaches, our method utilizes information in the solutions through math question back-translation, thereby enhancing the reliability of the augmented questions. We also create code-integrated solutions for the questions and use verification rationales to filter the solutions.

\section{Conclusion}

In this paper, we propose a coordinated pipeline consisting of \textit{Iterative Solution Augmentation} and \textit{Question Back-translation} to produce large-scale synthetic math questions, and \textit{Verification-Based Solution Filtering} to filter the generated code-integrated solutions. Combined, these three components effectively create new questions and ensure the reliability of the corresponding code-integrated solutions. Experiments show that MathGenieLM achieves superior performance across five math problem-solving benchmarks and on six different pretrained base models, offering insights into the development of math problem-solving models and providing hope for extension to other reasoning tasks.

\section*{Limitations}

Our method requires significant GPU resources, involving the full-parameter finetuning of large language models with up to 70B parameters. Therefore, it is crucial for future studies to explore ways to reduce the required resources. Another limitation is that our models cannot process images as input, and thus lack the ability to solve problems that involve images, as discussed in~\cite {lu2023mathvista}. Additionally, our models are constrained by a limited context length, having been finetuned with a context length of 4096. These limitations are significant and merit further investigation. 

\section*{Ethics Statement}

Our work, by enhancing the mathematical abilities of language models, can potentially contribute to the cause of math education. Still, our models can output untrue hallucinations, just like any language model. We have utilized various open-source models such as LLaMA-2, CodeLLaMA, Mistral, and Mixtral-8x7B, as well as open-source software such as Hugging Face and PyTorch. We adhere to the policies and licenses of these resources and acknowledge the role they have played in our work.

\section*{Acknowledgement}

This project is funded in part by National Key R\&D Program of China Project 2022ZD0161100, by the Centre for Perceptual and Interactive Intelligence (CPII) Ltd under the Innovation and Technology Commission (ITC)’s InnoHK, by General Research Fund of Hong Kong RGC Project 14204021. Hongsheng Li is a PI of CPII under the InnoHK. 

\bibliography{anthology,custom}
\bibliographystyle{acl_natbib}

\appendix



\section{Example of Iterative Solution Augmentation and Question Back-translation}
\label{app:backtranslation_examples}

\begin{figure*}[t]
    \centering
    \includegraphics[width=0.9\linewidth]{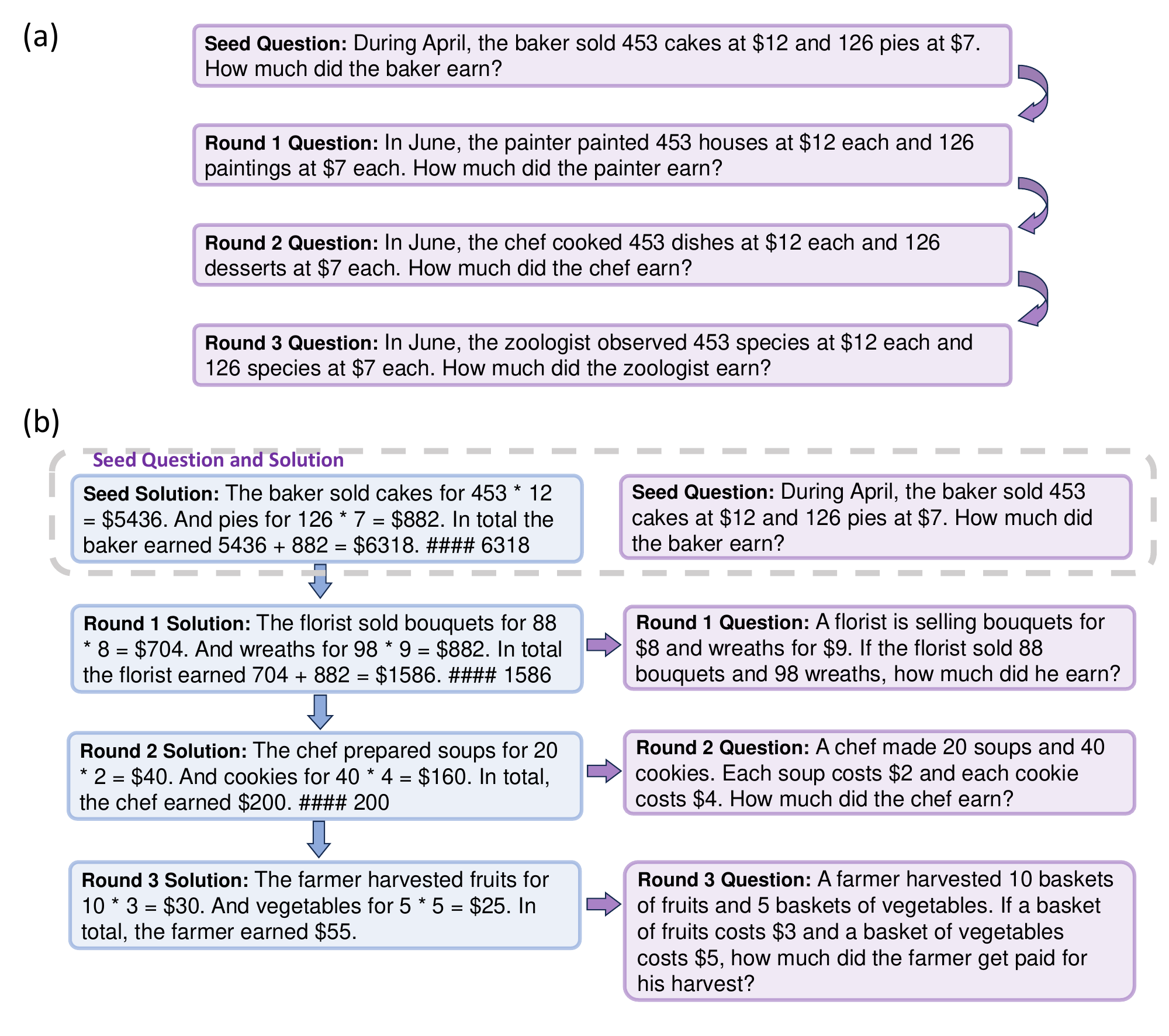}
    \caption{An example of three iterations of (a) Direct Question Augmentation and (b) Iterative Solution Augmentation and Question Back-translation. }
    
\label{fig:iteration_compare}
\end{figure*}

Fig.~\ref{fig:iteration_compare} (b) shows an example of three rounds of Iterative Solution Augmentation and Question Back-translation. The seed solution is iteratively augmented, and the augmented solutions are back-translated into new questions. Compared to Fig.~\ref{fig:iteration_compare} (a), where augmentation is conducted directly on the question, Iterative Solution Back-translation demonstrates greater diversity in question phrasing, as the original question is not directly provided to the model.

\begin{table*}[t]\fontsize{8.5}{9}\selectfont
\centering
\begin{tabularx}{\textwidth}{@{}*{2}{>{\centering\arraybackslash}X}|*{2}{>{\centering\arraybackslash\hsize=.85\hsize}X} | *{2}{>{\centering\arraybackslash\hsize=.85\hsize}X}  >{\centering\arraybackslash}X|*{1}{>{\centering\arraybackslash}X}@{}}
\midrule
&  & \multicolumn{2}{c|}{\textbf{In-Domain}} & \multicolumn{3}{c|}{\textbf{Out-of-Domain}}\\
& & \textbf{GSM8K} &   \textbf{MATH}   & \textbf{SVAMP}   & \textbf{Simuleq}        & \textbf{Mathematics}   & \textbf{Average}   \\
\midrule
\multicolumn{2}{l|}{\textbf{w/ iteration}}   & 80.5  &  45.1   & 83.3   & 79.4    & 71.8   & 72.0   \\
\midrule
\multicolumn{2}{l|}{\textbf{w/o iteration}}   & 78.7\tiny\color{green}({-1.8})  &  45.0\tiny\color{green}({-0.1})   & 82.9\tiny\color{green}({-0.4})   & 76.1\tiny\color{green}({-3.3})    & 70.9\tiny\color{green}({-0.9})   & 70.7\tiny\color{green}({-1.3})   \\
\midrule
\end{tabularx}
\caption{Comparison between with iteration or without iteration when conducting solution augmentation. The models are finetuned on Mistral 7B.}
\label{tab:iteration_ablation}
\end{table*}

\section{Analysis of Iterative Solution Augmentation}
\label{app:iteration_ablation}

To effectively demonstrate the impact of iteration on enhancing solution quality, we generated an equal number of augmented solutions without employing iteration, by directly augmenting new solutions from the original set. As illustrated in Table~\ref{tab:iteration_ablation}, the outcomes of the experiment conducted without iteration in solution augmentation were significantly inferior compared to those with iteration. This underscores the beneficial role of iteration in solution augmentation, primarily attributed to its potential to enhance the diversity of the solutions.

\section{Prompt of Verification}
\label{app:verify_input_prompt}

Tab.~\ref{tab:verification_format} presents the prompt format used in finetuning and generating code-integrated verification rationales.

\begin{table}[t]\fontsize{7.3}{2}\selectfont
\centering
\begin{tabularx}{\columnwidth}{X}
\toprule 
\textbf{Prompt} \\
\midrule
**Question**:

\{question\}

**Solution**:

\{solution\}

Above is a math problem and its solution. Please use code to verify the solution above.\\
\bottomrule 
\end{tabularx}
\caption{Prompt used for back-translation. \{question\} is replaced with a math question, while \{solution\} is replaced with the its code-integrated solution.}
\label{tab:verification_format}
\end{table}

\section{Analysis of Code-integrated Verification}
\label{sec:code_integrated_verification}

\begin{table*}[t!]\fontsize{8.5}{9}\selectfont
\centering
\begin{tabularx}{\textwidth}{@{}*{1}{>{\centering\arraybackslash}X}|*{5}{>{\centering\arraybackslash}X}|*{1}{>{\centering\arraybackslash}X}@{}}
\midrule
\textbf{Metrics}   & \textbf{GSM8K} &   \textbf{MATH}   & \textbf{SVAMP}   & \textbf{Simuleq}        & \textbf{Mathematics}   & \textbf{Average}   \\
\midrule
\textbf{Accuracy}   & 86.4  &  49.5   & 86.0   & 88.1    & 73.6   & 76.7   \\
\midrule
\textbf{Precision}   & 89.3\tiny\color{red}({+2.9})  &   64.5\tiny\color{red}({+15.0})   & 88.3\tiny\color{red}({+2.3})   & 90.8\tiny\color{red}({+2.7})   & 79.9\tiny\color{red}({+3.2})   & 82.6\tiny\color{red}({+5.9})  \\
\textbf{Recall}   & 98.5  &   94.7   & 97.4   & 98.2   & 97.1   &  97.2 \\
\midrule
\end{tabularx}
\caption{Accuracy, Precision and Recall of $M_{\rm code}$ with and without verification. Accuracy is the percent of solutions that are correct. Precision is the percent of solutions actually correct among the solutions verified to be correct. Recall is the percent of solutions verified to be correct among the solutions actually correct.}
\label{tab:verify_precision}
\end{table*}

To understand the reason behind the improved quality of the data, we quantify the ability of $M_{\rm code}$ to conduct code-integrated verification by testing it on the solutions generated by $M_{\rm code}$ across the five testing datasets. We utilize these testing datasets because they contain ground truth, which enables us to assess the actual correctness of the solutions. 

We define two metrics below to demonstrate the $M_{\rm code}$'s ability to verify its solutions: Precision and Recall.

$$\text{Precision}=\frac{\text{TP}}{\text{TP} + \text{FP}}$$
$$\text{Recall}=\frac{\text{TP}}{\text{TP} + \text{TN}}$$

TP represents the cases in which the verification proves the solution correct and the solution's answer is actually correct. FP represents the cases in which the verification proves the solution correct, yet the solution's answer is actually wrong. TN represents the cases in which the verification proves the solution wrong, yet the solution's answer is actually correct. In short, Precision answers the question, “What proportion of verified TRUE answers are actually correct?”, while Recall answers the question, “What proportion of actual correct answers were verified TRUE?”. Given these definitions, Precision reflects the reliability of the retained code-integrated solutions, while Recall reflects the efficiency of the filtering step.

Tab.~\ref{tab:verify_precision} shows that Precision is significantly higher than Accuracy across all datasets, underscoring the effectiveness and generalization capabilities of code-integrated verification.

\section{Finetuning Details}
\label{app:finetune_detail}

In this work, we finetune all models using the HuggingFace library. We employ a cosine weight scheduler with a learning rate of $2e^{-5}$, designating the first 50 steps as warm-up steps. All models are optimized using AdamW~\citep{loshchilov2017decoupled} with a batch size of 64. The 70B and 34B models are fine-tuned on 32 NVIDIA A800 80GB GPUs. Mistral-8x7B is fine-tuned on 16 NVIDIA A800 80GB GPUs, while the 7B, 13B, and 20B models are all fine-tuned on 8 NVIDIA A800 80GB GPUs.

\section{Analysis of Verification Rounds in Verified Inference}
\label{app:verification_rounds}

\begin{table*}[t!]\fontsize{8.5}{9}\selectfont
\centering
\begin{tabularx}{\textwidth}{@{}*{2}{>{\centering\arraybackslash}X}|*{5}{>{\centering\arraybackslash}X}|*{1}{>{\centering\arraybackslash}X}@{}}
\midrule
&  & \textbf{GSM8K} &   \textbf{MATH}   & \textbf{SVAMP}   & \textbf{Simuleq}        & \textbf{Mathematics}   & \textbf{Average}   \\
\midrule
\multirow{2}{*}{\textbf{Baseline}}  & \textbf{Accuracy}   & 88.4   & 51.2   &  87.7   & 89.1        & 76.0   & 78.5  \\
& \textbf{N$\times$}   & \textcolor{gray}{1$\times$}  &   \textcolor{gray}{1$\times$}   & \textcolor{gray}{1$\times$}   & \textcolor{gray}{1$\times$}   & \textcolor{gray}{1$\times$}   & \textcolor{gray}{1$\times$} \\
\midrule
\multirow{2}{*}{\textbf{Verify (k=1)}}  & \textbf{Accuracy}   & 88.4  &  54.2   & 88.4   & 90.9    & 79.2   & 80.2  \\
& \textbf{N$\times$}   & \textcolor{gray}{2.0$\times$}  &   \textcolor{gray}{2.3$\times$}   & \textcolor{gray}{2.0$\times$}   & \textcolor{gray}{2.0$\times$}   & \textcolor{gray}{2.1$\times$}   & \textcolor{gray}{2.1$\times$}  \\
\midrule
\multirow{2}{*}{\textbf{Verify (k=2)}}  & \textbf{Accuracy}   & 88.6  &  55.8  & 88.7   & 91.2    & 81.1   & 81.1  \\
& \textbf{N$\times$}   & \textcolor{gray}{2.1$\times$}  &   \textcolor{gray}{2.8$\times$}   & \textcolor{gray}{2.1$\times$}   & \textcolor{gray}{2.1$\times$}   & \textcolor{gray}{2.3$\times$}   & \textcolor{gray}{2.3$\times$}  \\
\midrule
\multirow{2}{*}{\textbf{Verify (k=3)}}  & \textbf{Accuracy}   & 88.6  &  56.3   & 88.8   & 91.2    & 81.3   & 81.2  \\
& \textbf{N$\times$}   & \textcolor{gray}{2.1$\times$}  &   \textcolor{gray}{3.1$\times$}   & \textcolor{gray}{2.1$\times$}   & \textcolor{gray}{2.1$\times$}   & \textcolor{gray}{2.3$\times$}   & \textcolor{gray}{2.3$\times$}  \\
\midrule
\multirow{2}{*}{\textbf{Verify (k=4)}}  & \textbf{Accuracy}   & 88.6  &  56.3   & 89.0   & 91.4    & 81.4   & 81.3  \\
& \textbf{N$\times$}   & \textcolor{gray}{2.1$\times$}  &   \textcolor{gray}{3.3$\times$}   & \textcolor{gray}{2.1$\times$}   & \textcolor{gray}{2.1$\times$}   & \textcolor{gray}{2.4$\times$}   & \textcolor{gray}{2.4$\times$}  \\
\midrule
\multirow{2}{*}{\textbf{Verify (k=5)}}  & \textbf{Accuracy}   & 88.6  &  56.5   & 88.9   & 91.4   & 81.4   & 81.4  \\
& \textbf{N$\times$}   & \textcolor{gray}{2.1$\times$}  &   \textcolor{gray}{3.4$\times$}   & \textcolor{gray}{2.1$\times$}   & \textcolor{gray}{2.1$\times$}   & \textcolor{gray}{2.4$\times$}   & \textcolor{gray}{2.4$\times$}  \\
\midrule
\multirow{2}{*}{\textbf{Verify (k=6)}}  & \textbf{Accuracy}   & 88.6  &  56.8   & 88.9   & 91.4    & 81.4   & 81.4  \\
& \textbf{N$\times$}   & \textcolor{gray}{2.1$\times$}  &   \textcolor{gray}{3.6$\times$}   & \textcolor{gray}{2.1$\times$}   & \textcolor{gray}{2.1$\times$}   & \textcolor{gray}{2.4$\times$}   & \textcolor{gray}{2.5$\times$}  \\
\midrule
\multirow{2}{*}{\textbf{Verify (k=7)}}  & \textbf{Accuracy}   & 88.6  &  56.8   & 88.9   & 91.4   & 81.4   & 81.4  \\
& \textbf{N$\times$}   & \textcolor{gray}{2.1$\times$}  &   \textcolor{gray}{3.7$\times$}   & \textcolor{gray}{2.1$\times$}   & \textcolor{gray}{2.1$\times$}   & \textcolor{gray}{2.4$\times$}   & \textcolor{gray}{2.5$\times$}  \\
\midrule
\multirow{2}{*}{\textbf{Verify (k=8)}}  & \textbf{Accuracy}   & 88.6  &  57.1   & 88.9   & 91.4    & 81.4   & 81.5  \\
& \textbf{N$\times$}   & \textcolor{gray}{2.1$\times$}  &   \textcolor{gray}{3.7$\times$}   & \textcolor{gray}{2.1$\times$}   & \textcolor{gray}{2.1$\times$}   & \textcolor{gray}{2.4$\times$}   & \textcolor{gray}{2.5$\times$}  \\
\midrule
\multirow{2}{*}{\textbf{Verify (k=9)}}  & \textbf{Accuracy}   & 88.6  &  57.1   & 88.9   & 91.4   & 81.4   & 81.5  \\
& \textbf{N$\times$}   & \textcolor{gray}{2.1$\times$}  &   \textcolor{gray}{3.8$\times$}   & \textcolor{gray}{2.1$\times$}   & \textcolor{gray}{2.1$\times$}   & \textcolor{gray}{2.4$\times$}   & \textcolor{gray}{2.5$\times$}  \\
\midrule
\end{tabularx}
\caption{Results of MathGenieLM-Llama-2-70B with varying numbers of verification rounds during inference. Here, ``k'' represents the maximum number of verification rounds.}
\label{tab:different_verified_inference_rounds}
\end{table*}

In verified inference, we verify the solutions of the test questions and re-solve only those questions whose solutions are verified as incorrect. Consequently, the number of questions that need solving decreases with each round. Theoretically, this process can continue until all questions have solutions verified as correct. However, in practice, verifying too many rounds can lead to additional costs without any improvement in accuracy, as some questions may be beyond the model's problem-solving and verification capabilities. To determine the trade-off between cost and accuracy, we increased the number of verification rounds to 9. The results are shown in Tab.~\ref{tab:different_verified_inference_rounds}. As can be seen, the increase in average accuracy becomes small after 2 rounds of verification.

\section{Examples of Code-Integrated Verification Rationales}
\label{app:verify_examples}
Two examples of code-integrated verification rationales are presented in Tab.~\ref{tab:example_verify_correct} and Tab.~\ref{tab:example_verify_wrong}. In Tab.~\ref{tab:example_verify_correct}, the solution is verified as correct by using the answer to calculate the condition and comparing it with the actual condition. In Tab.~\ref{tab:example_verify_wrong}, the solution is verified as incorrect by solving the question through an alternative method and comparing the answers.

\begin{table*}[t!]
    \centering
    \begin{tabular}{p{0.95\textwidth}}
    \toprule
\textbf{Input}:\\
**Question**:\\
     Express $0.\overline{54}$ as a fraction in lowest terms.\\
     **Solution**:\\
     To express the repeating decimal $0.\overline{54}$ as a fraction in lowest terms, we can use algebraic manipulation.\\
     Let $x = 0.\overline{54}$.\\
     Step 1: Multiply $x$ by 100 to shift the decimal point. \\
     ......\\
     Therefore, $0.\overline{54}$ can be expressed as $\frac{6}{11}$ in lowest terms.\\
     Above is a math problem and its solution. Please use code to verify the solution above. \\
     
    \midrule
\textbf{Output}:\\
Now, let's verify the answer by converting $\frac{6}{11}$ back to a decimal.
     \lstset{style=mystyle}
     \begin{lstlisting}[language=Python]
# Convert the fraction 6/11 to a decimal
decimal_value = float(simplified_fraction)

decimal_value
     \end{lstlisting} 
Result:\\
0.5454545454545454\\
The decimal representation of $\frac{6}{11}$ is approximately $0.5454545454545454$, which matches the repeating decimal $0.\overline{54}$\\
Thus, our answer is verified and correct.\\
     \bottomrule
    \end{tabular}
    \caption{An example of code-integrated verification rationale in which the solution is verified as correct.}
    \label{tab:example_verify_correct}
\end{table*}

\begin{table*}[t!]
    \centering
    \begin{tabular}{p{0.95\textwidth}}
    \toprule
\textbf{Input}:\\
**Question**:\\
What is $2121_3 - 212_3?$ Express your answer in base 3.\\
**Solution**:\\
Alright, let's solve this step by step.\\
To subtract these numbers, we'll start from the rightmost digit (least significant) and move to the left, just like we do in base 10 subtraction.\\
......\\
The result of $2121_3 - 212_3$ is $\boxed{222_3}$.\\
Above is a math problem and its solution. Please use code to verify the solution above.\\
    \midrule
\textbf{Output}:\\
Now, let's verify our answer by converting both numbers to base 10, performing the subtraction, and then converting the result back to base 3 to see if it matches our answer.
     \lstset{style=mystyle}
     \begin{lstlisting}[language=Python]
# Convert the numbers from base 3 to base 10
def base3_to_base10(num_str):
    num_str = num_str[::-1]  # reverse the string for easy processing
    base10_value = 0
    for i, digit in enumerate(num_str):
        base10_value += int(digit) * (3 ** i)
    ......
    
# Convert 2121_3 and 212_3 to base 10
num1_base10 = base3_to_base10("2121")
num2_base10 = base3_to_base10("212")

# Perform the subtraction in base 10
result_base10 = num1_base10 - num2_base10

# Convert the result back to base 3
result_base3 = base10_to_base3(result_base10)

result_base3
     \end{lstlisting} 
Result:\\
1202\\
It seems there was an error in our calculations. The correct result for $2121_3 - 212_3$ is $1202_3$, not $222_3$. I apologize for the oversight.
\\
     \bottomrule
    \end{tabular}
    \caption{An example of code-integrated verification rationale in which the solution is verified as wrong.}
    \label{tab:example_verify_wrong}
\end{table*}

\end{document}